\pdfoutput=1

\documentclass[11pt]{article}

\usepackage[]{emnlp2022}

\usepackage{times}
\usepackage{latexsym}

\usepackage[T1]{fontenc}

\usepackage[utf8]{inputenc}

\usepackage{microtype}

\usepackage{amsmath}
\usepackage{amsfonts}
\usepackage{arydshln}
\usepackage{booktabs}
\usepackage{bbding}
\usepackage{caption}
\usepackage{cprotect}
\usepackage{draftwatermark}
\usepackage{enumitem}
\usepackage{graphicx}
\usepackage{hyperref}
\usepackage{multicol}
\usepackage{multirow}
\usepackage{natbib}
\usepackage{paralist}
\usepackage{soul}
\usepackage{subcaption}
\usepackage[normalem]{ulem}
\usepackage{xspace}

\SetWatermarkText{}
\SetWatermarkScale{0.25}

\sethlcolor{RubineRed}
\newcommand{\fignum}[1]{\textcolor{white}{\hl{{\fontfamily{lmtt}\selectfont{#1}}}}}

\newcommand{\post}[1]{{\fontfamily{lmss}\selectfont{#1} \vspace{.25em}}}
\newcommand{\negative}{\textcolor{Red}{$\mathbf{-}$}}
\newcommand{\positive}{\textcolor{ForestGreen}{$\mathbf{\bigoplus}$}}
\newcommand{\neutral}{\textcolor{Cerulean}{{\fontfamily{bch}\selectfont{{\textbf{??}}}}}}
\newcommand{\conc}[1]{{\emph{\textcolor{NavyBlue}{#1}}}}

\newcommand{\texto}[1]{{\fontfamily{bch}\selectfont{{#1}}}}

\newcommand{\neurologic}{\texto{NeuroLogic}\xspace}

\newcommand{\neurocounterfactuals}{\texto{NeuroCounterfactuals}\xspace}
\newcommand{\neurocounterfactual}{\texto{NeuroCounterfactual}\xspace}

\newcommand{\ncfs}{\texto{NeuroCFs}\xspace}
\newcommand{\ncfsmini}{\texto{NeuroCFs}\xspace}
\newcommand{\ncf}{\texto{NeuroCF}\xspace}

\newcommand{\ncfuni}{\texto{NeuroCFs-1g}\xspace}
\newcommand{\ncfneu}{\texto{NeuroCFs-np}\xspace}

\newcommand{\draftonly}[1]{#1}
\renewcommand{\draftonly}[1]{}

\newcommand{\draftcomment}[1]{\draftonly{#1}}
\newcommand{\swabha}[1]{\draftcomment{\textcolor{orange}{\small [#1]$_{{SS}}$}}}
\newcommand{\phillip}[1]{\draftcomment{\textcolor{blue}{\small [#1]$_{{PH}}$}}}

\title{\neurocounterfactuals:\\ Beyond Minimal-Edit Counterfactuals for Richer Data Augmentation}

\newcommand{\aspace}{\hspace{1em}}
\newcommand{\uw}{$^{\heartsuit}$}
\newcommand{\intel}{$^{\diamondsuit}$}
\newcommand{\aiTwo}{$^{\clubsuit}$}
\newcommand{\usc}{$^{\spadesuit}$}
\author{
    Phillip Howard\intel\aspace 
    Gadi Singer\intel\aspace
    Vasudev Lal\intel \aspace 
    Yejin Choi\uw\aiTwo \aspace 
    Swabha Swayamdipta\aiTwo\usc\aspace \\
    \intel Intel Labs \aspace \aiTwo Allen Institute for AI \aspace \usc University of Southern California\\
    \uw Paul G.\ Allen School of Computer Science \& Engineering, University of Washington \\
    \texttt{phillip.r.howard@intel.com}
}

\begin{document}
\maketitle

\begin{abstract}
While counterfactual data augmentation offers a promising step towards robust generalization in natural language processing, producing a set of counterfactuals that offer valuable inductive bias for models remains a challenge. Most existing approaches for producing counterfactuals, manual or automated, rely on small perturbations via minimal edits, resulting in simplistic changes. We introduce \neurocounterfactuals, designed as loose counterfactuals, allowing for larger edits which result in naturalistic generations containing linguistic diversity, while still bearing similarity to the original document. Our novel generative approach bridges the benefits of constrained decoding, with those of language model adaptation for sentiment steering. Training data augmentation with our generations results in both in-domain and out-of-domain improvements for sentiment classification, outperforming even manually curated counterfactuals, under select settings. We further present detailed analyses to show the advantages of \neurocounterfactuals over approaches involving simple, minimal edits.
\end{abstract}
\section{Introduction}
\label{sec:introduction}

\begin{figure}[t!]
     \centering
         \includegraphics[width=0.95\columnwidth]{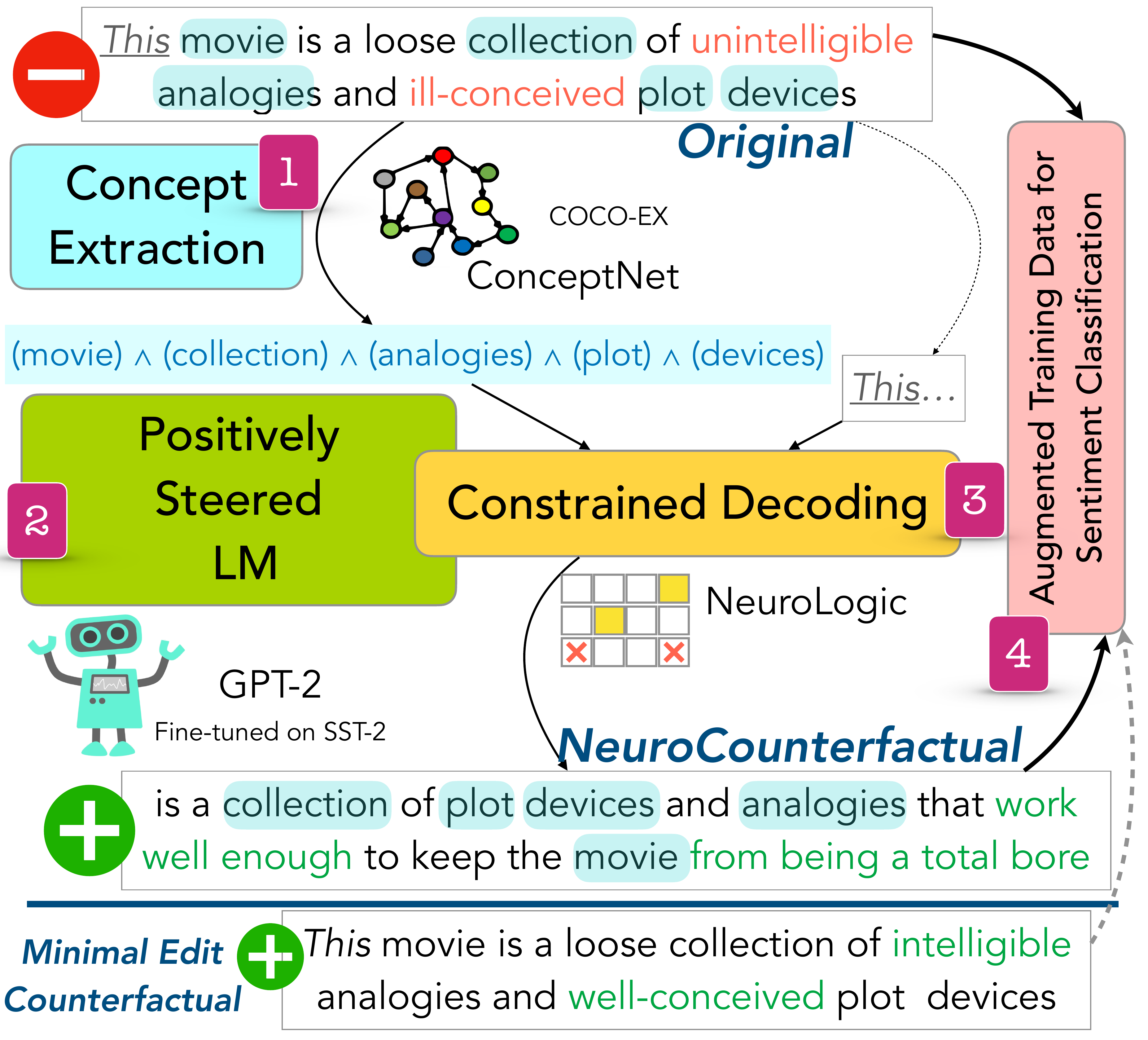}
         \caption{Illustration of our approach.
         \fignum{1} We extract tokens from an Original (negative) movie review that evoke concepts from ConceptNet (\S\ref{sec:constraints}).
         \fignum{2} We use a GPT-2 model adapted to only reviews with the opposite (positive) polarity as a sentiment steer (\S\ref{sec:lm-steering}).
         \fignum{3} Finally, to ensure that the generation is similar to the original, we use \neurologic, a constrained decoding approach (\S\ref{sec:neurologic}; \citealp{lu2021neurologic}), where the constraints are extracted tokens from \fignum{1}.
         This results in \neurocounterfactuals, which are loose counterfactuals of the original, but are more naturalistic (\S\ref{sec:analysis}; Tab.~\ref{tab:qualitative}), compared to minimal edit counterfactuals (bottom).
         \fignum{4} When used to augment training data for sentiment classification, our generations are valuable for OOD generalization (\S\ref{sec:experiments}).
         }
         \label{fig:neurocounterfactuals}
\end{figure}

Despite the enormous successes in natural language processing, out-of-domain (OOD) generalization still poses a challenge for even the most powerful models, which achieve remarkable performance in domain \cite{recht2019imagenet,torralba2011unbiased}.
This can be attributed to the models' reliance on spurious biases \cite{geirhos2020shortcut, mccoy-etal-2019-right, gururangan-etal-2018-annotation}, i.e. features which co-occur with the ground truth without any causal dependence \cite{simon1954spurious}.
Adopting methods from causal inference \cite{pearl2009causality,Feder2022Causal}, training data augmentation with counterfactuals (CFs) has been proposed for NLP as one potential solution \cite{levesque2012winograd,kaushik2019learning,kaushik2021explaining}.
Counterfactuals are designed to study the change in a response variable (e.g., the target label), following an intervention (e.g., altering a causal feature), typically in the form of edits to the input text \cite{khashabi2020bang,andreas-2020-good}.
Training data augmentation with counterfactuals can thus provide strong inductive biases to help with robustness against spurious biases, resulting in improved OOD generalization \cite{vig2020causal,eisenstein2022informativeness}. 

However, designing the appropriate interventions to produce counterfactuals can be challenging.
Indeed, most counterfactuals are produced via basic edits to the input text, either manually \cite{gardner2020evaluating,kaushik2019learning} or automatically \cite{yang2021exploring,wang2021robustness,wu2021polyjuice}, such that the target label changes. 
These minimal edits are made via substitution, insertion or deletion of tokens in the original sentence, resulting in simplistic generations, which are often unrealistic and lack linguistic diversity.\footnote{
For instance, the minimal edit counterfactual in \autoref{fig:neurocounterfactuals} contains the phrase ``\textit{loose collection of intelligible analogies}'', a somewhat unnatural construction for a positive movie review.}
As a result, counterfactuals via minimum edits often fail to provide adequate inductive biases to promote robustness \cite{khashabi2020bang,huang2020counterfactually,joshi2022investigation}.

In this paper, we investigate the potential of more realistic and creative counterfactuals, which go beyond simple token-level edits, towards improving robust generalization.
While allowing larger edits reduces proximity to the original sentence, we believe that this is a worthwhile trade-off for more realistic and creative counterfactuals, which offer greater flexibility in sentiment steering, increasing the likelihood that the counterfactual possesses the desired label.
We propose a novel approach that can generate diverse counterfactuals via concept-controlled text generation, illustrated in \autoref{fig:neurocounterfactuals}.
In particular, our approach combines 
the benefits of domain adaptive pretraining \cite{gururangan2020dont} for soft steering of the target label \cite{liu2021dexperts}, 
with those of \neurologic decoding \cite{lu2021neurologic}, an unsupervised, inference-time algorithm that generates fluent text while strictly satisfying complex lexical constraints. 
As constraints, we use tokens that evoke salient concepts derived from ConceptNet \cite{speer2017conceptnet}.
Our resulting generations, called \neurocounterfactuals\footnote{\ncfs, for short.}, 
provide loose counterfactuals to the original, while demonstrating nuanced linguistic alterations to change the target label (\S\ref{sec:methodology}).

Compared to minimal-edit counterfactuals, our counterfactuals are more natural and linguistically diverse, resulting in syntactic, semantic and pragmatic changes which alter the label while preserving relevance to the original concepts (\autoref{tab:qualitative}). 
On experiments with training data augmentation for sentiment classification, our approach achieves better performance compared to competitive baselines using minimal edit counterfactuals (\S\ref{sec:experiments}).
Our performance even matches baselines using human-annotated counterfactuals, on some settings, while avoiding the cost of human annotation.
While \ncfs are designed to be loose counterfactuals, our detailed analyses show that it is still important to augment training data with examples possessing a moderately high degree of similarity with the original examples (\S\ref{sec:analysis}).
When the ultimate goal is improving robust generalization, we show that going beyond minimal edit counterfactuals can result in richer data augmentation.\footnote{
Our code and data are available at \url{https://github.com/IntelLabs/NeuroCounterfactuals}}
\section{\neurocounterfactuals}
\label{sec:methodology}

\begin{table*}[ht]
\small
\resizebox{\textwidth}{!}{
\begin{tabular}{@{}p{1.9cm}p{0.5cm}p{16.5cm}@{}}
\toprule
\textbf{Source} & \textbf{Label} & \multicolumn{1}{c}{\textbf{Review}} \\ 
\midrule
Original & \negative 
& \post{But this \conc{film} decided to throw away the \conc{talents} of the \conc{people} involved in a simpering \conc{version} so watered down from the \conc{source material} that it's amazing they had the \conc{guts} to call it Wuthering \conc{Heights} at all.} \\
W\&C. &  \negative
& \post{But this film decided to throw away the talents of the people involved in a simpering version so watered down from the source material that it\underline{ s unimpressive} they had the guts to call it wuthering heights at all} \\
Y.et al. & \negative
& \post{But this film decided to throw away the talents of the people involved in a simpering version \sout{so} watered down from the source material that it's amazing they had the guts to call it wuthering heights at all.}  \\
\ncfuni &  \positive
& \post{But the \conc{film guts} its \conc{source material}, and it does so with a version of the \conc{heights} of artistry that \conc{people} have come to expect from the \conc{talents} of jean renoir.}  \\
\ncfneu & \positive
& \post{But this \conc{film} decided to take the \conc{talents} of the \conc{source material} and make them its own, and it's a \conc{guts}ier \conc{version} of the \conc{people} we know and love from the \conc{heights}.}  \\
\midrule
Original & \positive  
& \post{Unfortunately I had to rent a \conc{Dreamcast} to play it, but even though I did beat it I can't wait to buy it for \conc{PS2}.}\\
W\&C. & \positive
& \post{\underline{Fortunately} i had to rent a dreamcast to play it but even though i did beat it i can t wait to buy it for ps2}  \\
Y.et al. & \neutral
& \post{Unfortunately i had to rent a dreamcast to play it, but even though i did beat it i can't \sout{wait} to buy it for ps2.}  \\
\ncfuni & \negative  
& \post{Unfortunately it's not nearly as good as the \conc{dreamcast ps2} version.}  \\
\ncfneu & \negative
& \post{Unfortunately i had to rent a \conc{dreamcast} to play it but even though i did beat it i can't recommend it for \conc{ps2} or xbox.}  \\
\bottomrule
\end{tabular}
}
\caption{Comparison of IMDB-S train examples (Original) with generated counterfactuals from different approaches:  W\&C. \cite{wang2021robustness}, Y.et al. \cite{yang2021exploring}, and our \ncf variants, designed to flip the target label. 
The sentiment labels for the counterfactuals can be \positive~(positive), \negative~ (negative), or \neutral
~(unclear), as assessed by authors of this work. 
For the baselines, \underline{substitutions and insertions} are underlined, ignoring punctuation and capitalization, and \sout{deletions} are struck out.
\ncfs result in more complex changes to the original, and are more successful in steering the sentiment for label flipping; minimal edits are at times unable to result in meaningful changes to the sentiment, and result in reduced grammaticality.
Concepts in the original sentence that were used as constraints to generate \ncfsmini are \post{\conc{in blue italics}}.
Also see App~\S\ref{app:more-qualitative}; Tab.~\ref{tab:qualitative-more}.
}
\label{tab:qualitative}
\end{table*}

We describe our methodology for automatic generation of loose counterfactuals, \ncfs, for sentiment classification.
The key idea underlying our approach is the need for retention of concepts to ensure content similarity to the original text, while steering the sentiment to the opposite polarity.
Our method, illustrated in \autoref{fig:neurocounterfactuals}, combines a concept-constrained decoding strategy with a sentiment-steered language model.
First, we detail our approach for extracting the salient concepts from a document (\S\ref{sec:constraints}).
Next, we discuss language model adaptation to produce sentiment-steered LMs (\S\ref{sec:lm-steering}).
Finally, we provide an overview of the \neurologic decoding algorithm for controlled text generation, and how it can be adapted for the task of generating sentiment counterfactuals (\S\ref{sec:neurologic}).


\subsection{Extracting Salient Concepts}
\label{sec:constraints}

Our first step constitutes extraction of concepts from the original document, which can be used to reconstruct its content, when used as constraints during decoding (\S\ref{sec:neurologic}). 
Specifically, we aim to identify a set of constraints which will require the counterfactual to be similar in content to the original sentence while still allowing the generation to be steered towards the opposite polarity. 
Using extracted concepts as constraints achieves this because the concepts consist of the content-bearing noun phrases as opposed to the sentiment-bearing adjectives.
For example, in the original sentence from \autoref{fig:neurocounterfactuals}, we seek to constrain our generated counterfactual to contain concept-oriented phrases, such as \textit{``movie''}, \textit{``analogy''}, and \textit{``plot devices''} without explicitly requiring the presence of other tokens which may indicate the sentiment (e.g., \textit{``unintelligible''}, \textit{``ill-conceived''}). 

We achieve this mapping via linking tokens and phrases in the document to nodes in the ConceptNet knowledge graph \cite{speer2017conceptnet}, thus evoking salient concepts.
Nodes in ConceptNet are represented as non-canonicalized, free-form text. 
To this end, we use \textsc{COCO-EX} \cite{becker2021coco}, a ConceptNet entity linking tool. 
\textsc{COCO-EX} improves upon simple string-matching techniques which have been commonly used for ConceptNet entity linking in the past by selecting meaningful concepts and mapping them to a set of concept nodes based on relational information in the graph. 
Most extracted concepts correspond to nominal entities. 
Moreover, this mapping implicitly ensures that our extraction refrains from sentiment-bearing tokens and phrases.

We primarily use \textsc{COCO-EX} for its ability to identify meaningful concepts, but also explore the use of links to related concepts it provides in Section~\ref{sec:alternative-constraints}.
We also compare with a baseline using noun chunks as constraints in App~\ref{app:np-constraints}.


\subsection{Steering Sentiment via LM Adaptation}
\label{sec:lm-steering}

The second component for our method is a sentiment ``steer'', i.e. an autoregressive language model which has been trained or adapted via finetuning \cite{gururangan2020dont} exclusively on sentences with single (negative or positive) polarity.
Specifically, we use two steers for each sentiment label: one which models positive sentiment text, (denoted $p_{\theta}^{+}$), and another which models negative sentiment text, (denoted $p_{\theta}^{-}$), where $\theta$ indicates the parameters of the adapted language model.
In contrast to the hard predicate constraints over specific tokens as given by the extracted concepts in \S\ref{sec:constraints}, our selective use of steering LMs can be viewed as a softer type of constraint which biases the generations towards text containing the desired sentiment polarity \cite{liu2021dexperts}.
\swabha{FW: autoencoder, bart}

\subsection{Decoding with Conceptual Constraints}
\label{sec:neurologic}

Our method utilizes \neurologic Decoding 
\cite{lu2021neurologic}, a controlled text generation algorithm to generate fluent text satisfying a set of lexical constraints from a pretrained language model.
Given a series of predicates $D(\mathbf{a}, \mathbf{y})$ which are true iff $\mathbf{a}$ appears in the generated sequence $\mathbf{y}$, \neurologic accepts a set of \textit{clauses} $\{C_{i} \mid i \in 1, \cdots m\}$ consisting of one or more predicates specified in Conjunctive Normal Form (CNF):
\begin{equation*}
    \underbrace{(D_{1} \lor D_{2} \cdots \lor D_{i})}_\textrm{$C_1$} \land \cdots \land \underbrace{(D_{k} \lor D_{k+1} \cdots \lor D_{n})}_\textrm{$C_m$}
\end{equation*}
where each predicate $D_{i}$ is a positive constraint, $D(\mathbf{a}_i, \mathbf{y})$, which is satisfied (i.e., evaluates as true) if the subsequence $\mathbf{a}_i$ appears in the generated sequence $\mathbf{y}$. 

\neurologic employs a beam search approximation of an objective function which maximizes the probability of the generated sequence while penalizing deviations from the set of $m$ clauses:
\begin{equation}
    \hat{\mathbf{y}} = \text{arg} \max_{\mathbf{y} \in \mathcal{Y}} p_{\theta}(\mathbf{y} | \mathbf{x}) - \lambda \sum_{j=1}^{m} (1 - C_{j})
    \label{eq:objective}
\end{equation}
where $\lambda \gg 0$ penalizes deviations from the set of constraints. 
Candidates are scored at each stage $t$ of beam search according to their partial or full satisfaction of the constraints:
\begin{equation}
    f(\mathbf{y}_{\le t}) = \log p_{\theta} (\mathbf{y}_{\le t} | \mathbf{x}) + \lambda \max_{D(\mathbf{a}, \mathbf{y}_{\le t})} \frac{|\hat{\mathbf{a}}|}{|\mathbf{a}|}
    \label{eq:scoring}
\end{equation}
where $\hat{\mathbf{a}}$ represents a subsequence of $\mathbf{a}$ in the current generation and the maximum is taken over all unsatisfied constraints consisting of more than one token. 
This has the effect of preferring candidates which at least partially satisfy multi-token constraints; for example, a generated sequence $\mathbf{y}_{\le t} =$ {``The boy climbs an apple''} would be rewarded for partially satisfying the constraint $\mathbf{a} =$ {``apple tree''} via its subsequence $\hat{\mathbf{a}} =$ {``apple''}. 

Unlike the top-$k$ selection strategy used in traditional beam search, \neurologic performs pruning, grouping, and selection steps to identify the best candidates which satisfy the given constraints. 
Specifically, candidates which irreversibly violate one or more constraints are pruned, and the remaining candidates are grouped according to their number of satisfied clauses in order to encourage diversity. 
The best candidate within each group is then selected according to the scoring function in Equation~\ref{eq:scoring}.

Each word or phrase in the original example which is linked to a ConceptNet node (\S\ref{sec:constraints}) becomes a clause in our constraint set used with \neurologic. 
We allow each clause to be satisfied by the lowercase or capitalized form of the concept via an OR constraint.
For the example in \autoref{fig:neurocounterfactuals}, this constraint set would be specified in CNF as follows:
\begin{multline*}
\small
    (\textit{Movie} \lor \textit{movie}) \land (\textit{Plot Devices} \lor \textit{plot devices}) \land \\ \small (\textit{Collection} \lor \textit{collection}) \land (\textit{Analogies} \lor \textit{analogies})
\end{multline*}
\swabha{PHILLIP: do we also use capitalization as an OR constraint???}

Once the constraints have been identified in the original, we substitute the sentiment-steered LMs (\S\ref{sec:lm-steering}) into Equation~\ref{eq:objective}, corresponding to a polarity opposite to the original:
\vspace{-3mm}
\begin{equation}
    \hat{\mathbf{y}} = \text{arg} \max_{\mathbf{y} \in \mathcal{Y}} p_{\theta}^{i}(\mathbf{y} | \mathbf{x}) - \lambda \sum_{j=1}^{m} (1 - C_{j}).
    \label{eq:objectiveNew}
\end{equation}
Here, $p_{\theta}^{i} = p_{\theta}^{+}$ when we aim to generate a positive-sentiment example and $p_{\theta}^{i} = p_{\theta}^{-}$, for a negative-sentiment example.
The resulting generation, $\hat{\mathbf{y}}$, is a \textbf{\neurocounterfactual (\ncf)}.

In Eq.~\ref{eq:objectiveNew}, the generation is conditioned on $\mathbf{x}$, which indicates a prompt, comprising a prefix of the original input; we investigate two variants for $\mathbf{x}$.
When $\mathbf{x}$ is a unigram (\underline{1g}) comprising the \textit{first token} of the original input, we call the generations \textbf{\ncfuni}.
When $\mathbf{x}$ is the \textit{longest \underline{n}eutral \underline{p}refix} of the original input, we call the generations \textbf{\ncfneu}; these are slightly tighter \ncfs containing a greater portion of the original input. 
\autoref{tab:qualitative} provides examples showing the original sentence and our generated \ncfsmini, highlighting words in the original that were included in the concept-oriented constraint set for \neurologic decoding. 
\ncfsmini are not guaranteed to \emph{not} contain new concepts, beyond the specifications of the constraint set.
See App.~\S\ref{app:more-qualitative} for further examples.

\section{Data Augmentation with \ncfs}
\label{sec:experiments}

Our experiments compare \ncfs to CFs from minimal edit approaches, for augmentation of sentiment classification training data.

\subsection{Experimental Setup}
\label{sec:setup}

\paragraph{Sentiment Steer} 
Our positive and negative sentiment steers are based on a GPT-2 Large model \cite{Radford2019LanguageMA}, finetuned on (positive and negative, resp.) subsets of the Stanford Sentiment Treebank (SST-2; \citealp{socher2013recursive}) corpus, including train, test and validation splits.\footnote{
We use the \href{{https://github.com/alisawuffles/DExperts}}{sentiment experts} released by \citet{liu2021dexperts}.}

\paragraph{\neurologic}
For decoding with \neurologic, we use a beam size of 20, length penalty of 0.3, and an $n$-gram size of 2 for preventing repetitions.
We use $\beta = 1.25$ as the reward factor for in-progress constraint satisfaction and set the constraint satisfaction tolerance to 2.
Please refer to \citet{lu2021neurologic} for details on these hyperparameters.

For the generation of \ncfneu, we identify the longest neutral prefix of the original input.
As candidates, we consider all prefixes containing at least 4 tokens, such that the rest of the review contains at least one identified concept. 
We filter the longest candidate, predicted as neutral using an off-the-shelf 5-way sentiment classifier.\footnote{From \href{https://github.com/ShannonAI/Self_Explaining_Structures_Improve_NLP_Models}{ShannonAI}.}

Following prior work \cite{kaushik2019learning}, we generate \ncfs for a subset of movie reviews from the Internet Movie Database (IMDB; \citealp{pang2005seeing}), comprising 2440 examples randomly sampled and split into 70\% training, 10\% validation, and 20\% test partitions, for a sentiment classification task \cite{maas2011learning}.
We augment the training data of a sentence-level version of this dataset (\textbf{IMDB-S})\footnote{
Initial experiments with \neurologic decoding with full length IMDB paragraphs were prohibitively slow, which we circumvented by using the sentence-level version.
}, introduced by \citet{wang2021robustness}; see App.~\ref{sec:imdb-s} for details.

\begin{table*}[ht!]
\centering
\small
\resizebox{1\textwidth}{!}{%
\begin{tabular}{l r  c c c c c c c c}
\toprule
& & \multicolumn{3}{c}{\textbf{IMDB}}  & \multicolumn{2}{c}{\textbf{SST-2}} & \multicolumn{3}{c}{\textbf{Out-of-domain}}  \\
\cmidrule(lr){3-5}
\cmidrule(lr){6-7}
\cmidrule(lr){8-10}
\textbf{Source of CFs} & \textbf{$|D_\textrm{train}|$} &  \textbf{Test} &   \textbf{CF (K. et al.)} & \textbf{Cont.Sets} &\textbf{Test} & \textbf{PolyJuice CFs} & \textbf{Twitter} & \textbf{Yelp} & \textbf{Amazon} \\
\midrule
None & 8,173 & $93.22_{0.42}$ & $92.07_{1.04}$ & $86.85_{1.06}$ & $90.30_{0.97}$ & $84.74_{0.46}$ & $77.94_{1.72}$ & $94.71_{0.67}$ & $90.35_{1.03}$ \\
\midrule
\citealp{yang2021exploring} & 10,376 & $92.15_{0.79}$ & $91.99_{1.56}$ & $86.67_{1.46}$ & $89.46_{0.95}$ & $86.90_{0.57}$ & $76.37_{1.96}$ & $94.23_{0.59}$ & $89.97_{1.07}$\\
\citealp{wang2021robustness} & 10,744 & $\mathbf{92.88}_{0.45}$ & $94.03_{0.91}$ & $89.69_{0.87}$ & $89.26_{1.55}$ & $85.97_{0.69}$ & $77.09_{1.97}$ & $94.47_{0.61}$ & $90.88_{0.89}$\\
\ncfuni & 15,437 & $92.60_{0.59}$ & $93.36_{0.71}$ & $89.04_{1.02}$ & $\underline{\mathbf{92.63}}_{0.44}$ & $87.11_{0.52}$ & $\underline{77.98}_{1.22}$ & $\underline{\mathbf{95.01}}_{0.22}$ & $\underline{\mathbf{92.32}}_{0.51}$\\
\ncfneu & 12,905 & $92.66_{0.46}$ & $\underline{\mathbf{95.03}}_{0.47}$ & $\underline{\mathbf{90.85}}_{0.84}$ & $\underline{92.27}_{0.39}$ & $\underline{\mathbf{88.35}}_{0.41}$ & $\underline{\mathbf{78.80}}_{1.22}$ & $94.51_{0.87}$ & $\underline{92.24}_{0.71}$\\
\midrule
\HandPencilLeft~\citealp{kaushik2019learning} & 16,679 & $92.63_{0.48}$ & $97.34_{0.37}$ & $95.22_{0.45}$ & $89.73_{0.76}$ & $90.10_{0.29}$ & $81.28_{1.60}$ & $93.94_{0.52}$ & $91.96_{0.44}$\\
\bottomrule
\end{tabular}}
\caption{
Sentiment classification accuracies, comparing IMDB-S training data augmentation with \ncfs vs. other sources of counterfactuals. IMDB CF (K. et al.) and Cont.Sets refer to the human-authored counterfactuals \cite{kaushik2019learning} and contrast sets \cite{gardner2020evaluating}, respectively.
$|D_\textrm{train}|$ shows the total number of training instances, \textit{including} 8,173 original IMDB-S training examples.
Results report mean over 30 differnt random seeds, with s.d. as a subscript.
All models are based on the RoBERTa-base architecture.
Best results using auto-generated CFs for training are in boldface. Results for \ncfuni and \ncfneu are underlined when a one-tailed t-test indicates that their improvements over both \citealp{yang2021exploring} and \citealp{wang2021robustness} are statistically significant ($p \le 0.05)$. 
\HandPencilLeft~indicates manually created counterfactuals.
}
\label{tab:fullCtfSet}
\end{table*}

\subsection{Baselines: Other CF sources}
\label{sec:baselines}

We compare with multiple sentiment classification baselines employing counterfactuals for training data augmentation.
\citet{kaushik2019learning} crowdsource counterfactuals for IMDB, by soliciting minimal revisions to maintain coherence while flipping the sentiment, creating both a counterfactually augmented train as well as test dataset. 
We also consider two approaches that produce automatically generated counterfactuals. 
\citet{wang2021robustness} generate counterfactuals by automatically identifying causal words in the original example and substituting them with their antonyms, ensuring minimal edits. 
Similarly, \citet{yang2021exploring} automate counterfactual generation through the identification of causal terms which are either removed or replaced; they then filter candidates using MoverScore \cite{zhao2019moverscore} to ensure minimal edits were made to the original example.
For all the baselines above, we train on sentence-level IMDB reviews, as well as sentence-level variants of the counterfactuals.\footnote{App.~\ref{sec:dataset-details} provides further details on our datasets.}

\paragraph{Sentiment Classifier}
We compare several models, based on a RoBERTa-base architecture \cite{liu2019roberta}.
Each model is trained on a counterfactually augmented dataset, where the CFs are either obtained via baselines above, or via our approaches (\S\ref{sec:methodology}).
We additionally train a baseline only on the original IMDB-S training data, without any CFs.
Details on model training are provided in App.~\ref{sec:model-details}.

\paragraph{Evaluation}
We report classification accuracy on a combination of in-domain and out-of-domain test sets.
As in-domain test sets, we evaluate on the IMDB test set.
We also evaluate on CFs for IMDB, crowdsourced by \citet{kaushik2019learning}.
In addition, we evaluate on contrast sets \cite{gardner2020evaluating}, which are expert-annotated CFs for IMDB test data.
As another in-domain test set, we evaluate on the SST-2 movie reviews test set.\footnote{
While our sentiment steers are trained on SST-2 data, we use \neurologic decoding to obtain counterfactuals for IMDB, which we use to train our sentiment classifier. 
Hence, it is unlikely that the classifier is exposed to SST-2 directly.
}
\citet{wu2021polyjuice} produce task-agnostic, minimal edit counterfactuals with fine-grained semantic controls over different types of perturbations, followed by human labeling;  we also evaluate on these so-called Polyjuice CFs for SST-2 test.\footnote{We cannot compare with a baseline trained on Polyjuice CFs, as these are not available for IMDB, and would need human labeling.}
While SST-2 differs from IMDB in terms of word length and style, we nevertheless consider it in-domain for the purpose of our evaluations because both datasets are comprised of movie reviews.

For the OOD test sets, we consider the following binary sentiment classification datasets:
\begin{compactitem}
\item The \textbf{Amazon} dataset \cite{ni2019justifying} consists of consumer product reviews in the categories of software, fashion, appliances, beauty, magazines, and gift cards. 
\item The \textbf{Twitter} dataset \cite{rosenthal2017semeval} from SemEval-2017 Task 4 contains social media posts collected from Twitter. 
\item The \textbf{Yelp} dataset\footnote{\url{https://www.yelp.com/dataset}} contains consumer reviews originating from the Yelp dataset challenge.
\end{compactitem}

\begin{table*}[ht]
\centering
\small
\resizebox{1\textwidth}{!}{%
\begin{tabular}{c r l c c c c c c c c }
\toprule
& & & \multicolumn{3}{c}{\textbf{IMDB}} & \multicolumn{2}{c}{\textbf{SST-2}} & \multicolumn{3}{c}{\textbf{Out-of-domain}}  \\
\cmidrule(lr){4-6}
\cmidrule(lr){7-8}
\cmidrule(lr){9-11}
 & \textbf{$|D^\textrm{CF}_\textrm{train}|$} &  \textbf{Source of CFs} & \textbf{Test} & \textbf{CF (K. et al.)} & \textbf{Cont.Sets} & \textbf{Test} & \textbf{PolyJuice CFs}  & \textbf{Twitter} & \textbf{Yelp} & \textbf{Amazon} \\
\midrule
\multirow{6}{*}{\rotatebox[origin=c]{90}{\textbf{$|D_\textrm{train}| = 8,173$}}} & $0$ & None &  $93.22_{0.42}$ & $92.07_{1.04}$ & $86.85_{1.06}$ & $90.30_{0.97}$ & $84.74_{0.46}$ & $77.94_{1.72}$ & $94.71_{0.67}$ & $90.35_{1.03}$ \\
\cmidrule(lr){2-11}
 & \multirow{5}{*}{\rotatebox{0}{$2,203$}} & \citealp{yang2021exploring} & $91.68_{0.91}$ & $91.91_{1.65}$ & $86.69_{1.76}$ & $89.73_{1.05}$ & $87.24_{0.51}$ & $77.03_{2.20}$ & $93.22_{1.31}$ & $90.02_{1.20}$\\
& & \citealp{wang2021robustness} & $92.66_{0.52}$ & $94.17_{1.21}$ & $89.41_{1.50}$ & $89.15_{1.30}$ & $85.87_{0.53}$ & $77.62_{1.67}$ & $94.23_{0.70}$ & $90.98_{0.84}$ \\
& & \ncfuni & $92.58_{0.71}$ & $93.35_{0.86}$ & $88.20_{1.12}$ & $\underline{\mathbf{92.13}}_{0.60}$ & $86.63_{0.55}$ & $\underline{\mathbf{78.88}}_{1.37}$ & $\underline{\mathbf{94.93}}_{0.52}$ & $\underline{91.80}_{0.72}$ \\
& & \ncfneu &  $\underline{\mathbf{92.87}}_{0.46}$ & $\underline{\mathbf{94.75}}_{0.64}$ & $\underline{\mathbf{89.99}}_{0.94}$ & $\underline{92.04}_{0.70}$ & $\underline{\mathbf{87.64}}_{0.57}$ & $\underline{78.72}_{1.51}$ & $\underline{94.76}_{0.55}$ & $\underline{\mathbf{91.87}}_{0.73}$ \\
\cmidrule(lr){3-11}
& & \HandPencilLeft~\citealp{kaushik2019learning} & $93.09_{0.46}$ & $96.06_{0.37}$ & $92.81_{0.79}$ & $90.99_{0.82}$ & $88.48_{0.43}$ & $80.30_{1.60}$ & $94.52_{0.81}$ & $91.87_{0.89}$ \\
\bottomrule
\end{tabular}}
\caption{Results controlling for training data quantity ($|D_\textrm{train}|$), comparing different counterfactual data augmentaton approaches.
The first row shows a baseline trained \textit{without} CFs.
All other settings are identical to \autoref{tab:fullCtfSet}.
}
\label{tab:oodTestSameSize}
\end{table*}
\begin{table*}[ht]
\centering
\small
\begin{tabular}{l c c c c c}
\toprule
\textbf{Source of CFs}& \textbf{BLEU} &  \textbf{Levenshtein} & \textbf{MoverScore} & \textbf{Ppl} & \textbf{Distinct-2}\\
\midrule
\HandPencilLeft~\citealp{kaushik2019learning} &  0.74 & 20 & 0.70 & 19.3 & 0.49\\
\midrule
\citealp{yang2021exploring} & 0.80 & 8 & 0.81 & 29.1 & 0.56\\
\citealp{wang2021robustness} & 0.56 &  13 & 0.65 & 65.6 & 0.58\\
\ncfneu & 0.50 & 48 & 0.46 & 12.7 & 0.45\\
$\quad$ w/ concept-altered & 0.43 &  70 & 0.38 & 18.5 & 0.51\\
\ncfuni & 0.10 &  89 & 0.20 & 14.1 & 0.38\\
$\quad$ w/o constraints & 0.03 &  97 & 0.07 & 4.6 & 0.32\\
\bottomrule
\end{tabular}
\caption{Comparing fluency, diversity, and similarity of generated and human (\HandPencilLeft) CFs to the original, across various metrics.
\ncfs are loose counterfactuals by design, and are therefore farther away from the original sentence.
\phillip{Note that I added perplexity and dinstct-2 to this table, and broadened its scope beyond what it had before (which was just metrics relating the original example to the counterfactual)}\swabha{lgtm!}}
\label{tab:intrinsicEvaluation}
\end{table*}
\subsection{\ncfsmini for Train Data Augmentation}
\label{sec:training-daug}

\autoref{tab:fullCtfSet} shows our results.
\ncfs outperform alternative methods for automatic CF generation across every in-domain as well as OOD setting, including performance on CF test sets.
The only exception is IMDB test, where we match the performance of the best approach (up to standard deviation).
Across most CF and OOD test sets, the magnitude of our improvements is similar to or greater than the amount by which existing methods improve on the no-counterfactual baseline.
Furthermore, most of these improvements are statistically significant $(p \le 0.05)$ relative to the results of both \citet{yang2021exploring} and \citet{wang2021robustness}.
\ncfs even surpass the performance of augmentation with crowdsourced counterfactuals from \citet{kaushik2019learning} on most OOD settings.
However, training on manual CFs results in higher performance when tested on human-written CFs; this might be attributed to distributional similarities \cite{geirhos2020shortcut,koh2020wilds}.
Regardless, our performance is close enough, despite using fewer training instances while avoiding the significant cost of human annotation.

Both \ncf-variants have comparable performance, with the \ncfneu faring better on 4/8 benchmarks.
Consistent with prior work \cite{wang2021robustness}, we observe that training with CFs generally results in similar or slightly worse in-domain test performance on IMDB-Test, relative to training without CFs. 

Each source of CFs evaluated in \autoref{tab:fullCtfSet} produces different amounts of training data, $D_\textrm{train}$.
To control for training data quantity, we present results with downsampling the training data for uniformity across settings, in \autoref{tab:oodTestSameSize}.
Surprisingly, even lower amounts of \ncfs achieve the best performance compared to other methods of autogenerating CFs.
Notably, \ncfneu achieves statistically significant improvements over both \citet{yang2021exploring} and \citet{wang2021robustness} on every evaluated dataset.
These results demonstrate that the performance improvements achieved on OOD sets can be attributed to the quality of the \ncfs.
App.~\ref{app:slevel-eval} provides further results on sentence-level tests.

Table~\ref{tab:intrinsicEvaluation} compares our \ncfs and CFs from other sources, to the original, across three similarity metrics: BLEU ($n$-gram$=2$) \cite{papineni-etal-2002-bleu}, Levenshtein edit distance \cite{levenshtein1966binary}, and MoverScore \cite{zhao2019moverscore}.
Additionally, Table~\ref{tab:intrinsicEvaluation} provides the mean perplexity of generated counterfactuals as measured by GPT-J \cite{gpt-j} as well as the Distinct-2 diversity measure \cite{li2015diversity}. 
\ncfs are loose counterfactuals by design, and are therefore farther away from the original sentence; \ncfneu are tighter CFs compared to \ncfuni.
However, \ncfs have greater fluency (as evidenced by lower mean perplexity) and offer performance benefits over more similar CFs via minimal edit approaches (\autoref{tab:fullCtfSet}).
Moreover, more dissimilar variants, generated without constraints for generation (\S\ref{sec:analysis-constraints}), or with alternative concepts (\S\ref{sec:alternative-constraints}) also hurt performance.

\section{Analysing \ncfs}
\label{sec:analysis}

We present further analysis of \ncf properties, such as \ncfsmini size (\S\ref{sec:cad-size}), and similarity to the original (\S\ref{sec:diversity}), and also ablations of our method (\S\ref{sec:analysis-constraints}, \S\ref{sec:alternative-constraints}). 

\subsection{Impact of \ncf quantity}
\label{sec:cad-size}
In contrast to minimal edit approaches, our approach has the added advantage of producing more than a single \ncf for each original example, via \neurologic hyperparameter variation. 
We seek to investigate how the quantity of \ncfs for training data augmentation impacts OOD generalization.
To investigate the effect of size beyond results in \autoref{tab:fullCtfSet}, we generate more \ncfneu by varying the length penalty in \neurologic from 0.1 to 0.7 in increments of 0.2. 
Among these candidate counterfactuals for each original instance, we augment the training data with the generation with the lowest MoverScore to our initial \ncfneu.
This increases the quantity of \ncfneu from 4,732 to 7,489. 

\begin{figure}[h]
     \centering
         \includegraphics[width=0.95\columnwidth]{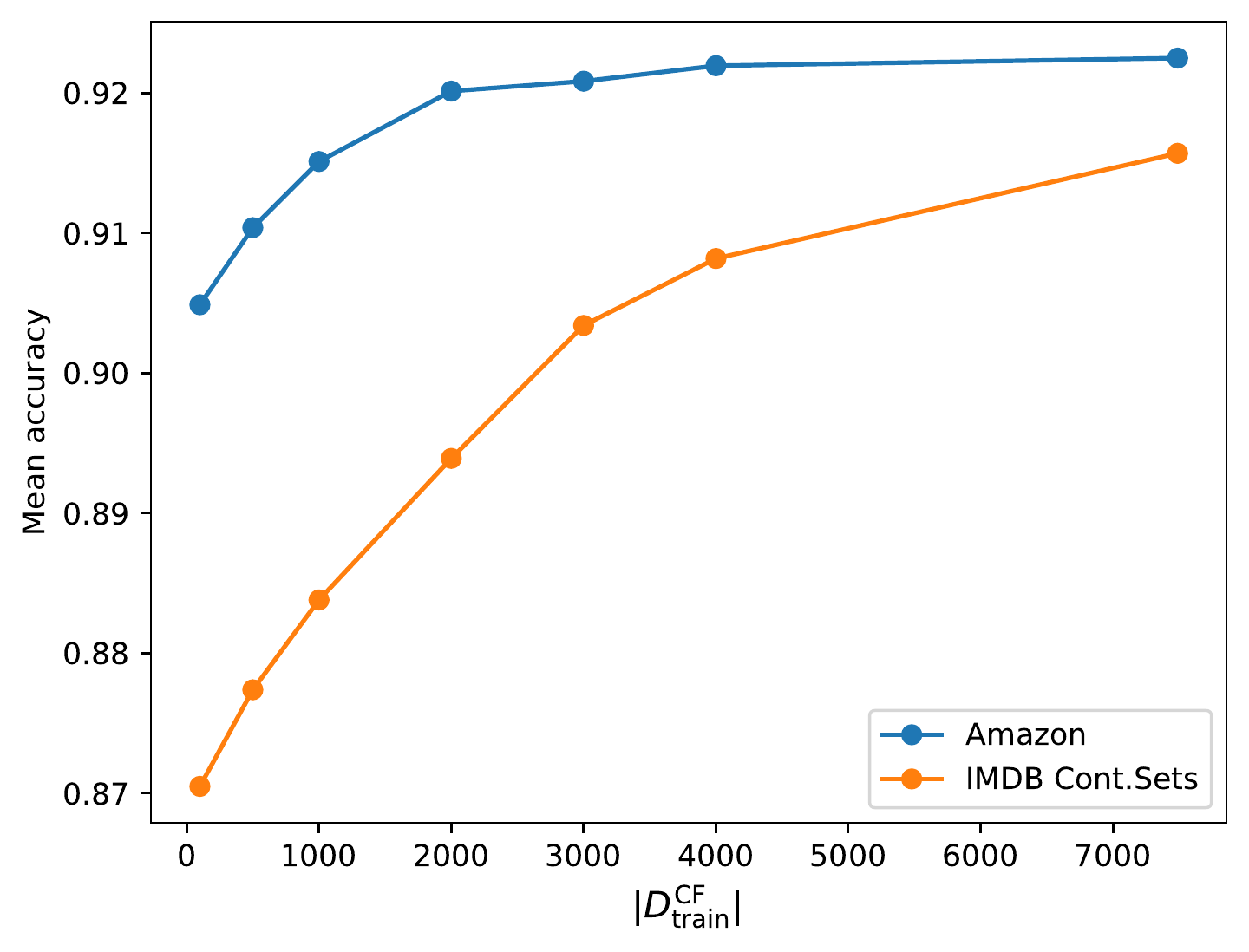}
         \caption{Increasing \ncf quantity for training data augmentation improves in-domain performance, while OOD generalization plateaus.
         } 
         \label{fig:cadSize}
\end{figure}
Results in Figure~\ref{fig:cadSize} show monotonic increase in accuracy on IMDB contrast sets \cite{gardner2020evaluating} with \ncfneu size. 
However, performance on the Amazon OOD set plateaus, suggesting overfitting to the IMDB domain; this echoes the findings of prior work on the efficacy of counterfactuals \cite{khashabi2020bang,huang2020counterfactually,joshi2022investigation}.




\subsection{Impact of \ncf Similarity}
\label{sec:diversity}
\begin{table*}[ht]
\centering
\resizebox{1\textwidth}{!}{%
\begin{tabular}{l c c c c c c c c c}
\toprule
&  & \multicolumn{3}{c}{\textbf{IMDB}} & \multicolumn{2}{c}{\textbf{SST-2}} & \multicolumn{3}{c}{\textbf{Out-of-domain}}  \\
\cmidrule(lr){3-5}
\cmidrule(lr){6-7}
\cmidrule(lr){8-10}
\textbf{\ncf-Variants} & \textbf{MoverScore} & \textbf{Test} & \textbf{CF (K. et al.)} & \textbf{Cont.Sets} &\textbf{Test} & \textbf{CF (PolyJuice)} & \textbf{Twitter} & \textbf{Yelp} & \textbf{Amazon} \\
\midrule
\ncfsmini$_{\textrm{loose}}$ & 0.114 & $\mathbf{92.50}_{0.59}$ & $93.31_{0.71}$ & $88.35_{0.71}$ & $\mathbf{92.26}_{0.56}$ & $86.56_{0.44}$ & $76.95_{1.62}$ & $\mathbf{95.01}_{0.42}$ & $91.51_{0.78}$\\
\ncfsmini$_{\textrm{tight}}$ & 0.373 & $92.24_{0.68}$ & $\mathbf{93.33}_{0.56}$ & $\mathbf{89.29}_{0.71}$ & $92.23_{0.55}$ & $\mathbf{86.80}_{0.41}$ & $\mathbf{77.73}_{1.22}$ & $94.93_{0.28}$ & $\mathbf{92.00}_{0.59}$\\
\bottomrule
\end{tabular}}
\caption{Impact of the similarity of a \ncf to the original. 
\ncfsmini$_{\textrm{loose}}$ are more dissimilar to the original, than \ncfsmini$_{\textrm{tight}}$, as given by the mean MoverScore.
Tighter \ncfs result in better performance. 
}
\label{tab:ctfSimilarity}
\end{table*}

We investigate the impact of the similarity of \ncfs to the original example on sentiment classification performance after augmentation.
From the \ncfs candidate set described in \S\ref{sec:cad-size}, we create two sets of alternative \ncfs for each instance: one with the lowest MoverScore (most dissimilar) w.r.t. the original (\ncfs$_{\textrm{loose}}$) and the other with the highest MoverScore (most similar; \ncfs$_{\textrm{tight}}$). 

Table~\ref{tab:ctfSimilarity} compares these two alternatives via classifier performance across our in-domain and out-of-domain tests. 
In general, we observe that tighter (i.e., more similar to the original sentence) counterfactuals improve generalization more when evaluated on counterfactual and contrast sets. 
They also improve out-of-domain generalization, with the exception of the Yelp dataset where both variants result in similar performance.
Tighter counterfactuals are more likely to break spurious correlations that help classifiers perform better on in-domain test sets, which may explain why \ncfs$_{\textrm{loose}}$ performs better on IMDB Test and SST Test.
While \ncfs are designed to be loose CFs, these results suggest that higher similarity between the original and its \ncf is still important for generalization.

\subsection{Impact of Constrained Decoding}
\label{sec:analysis-constraints}

\begin{figure}
     \centering
         \includegraphics[width=0.95\columnwidth]{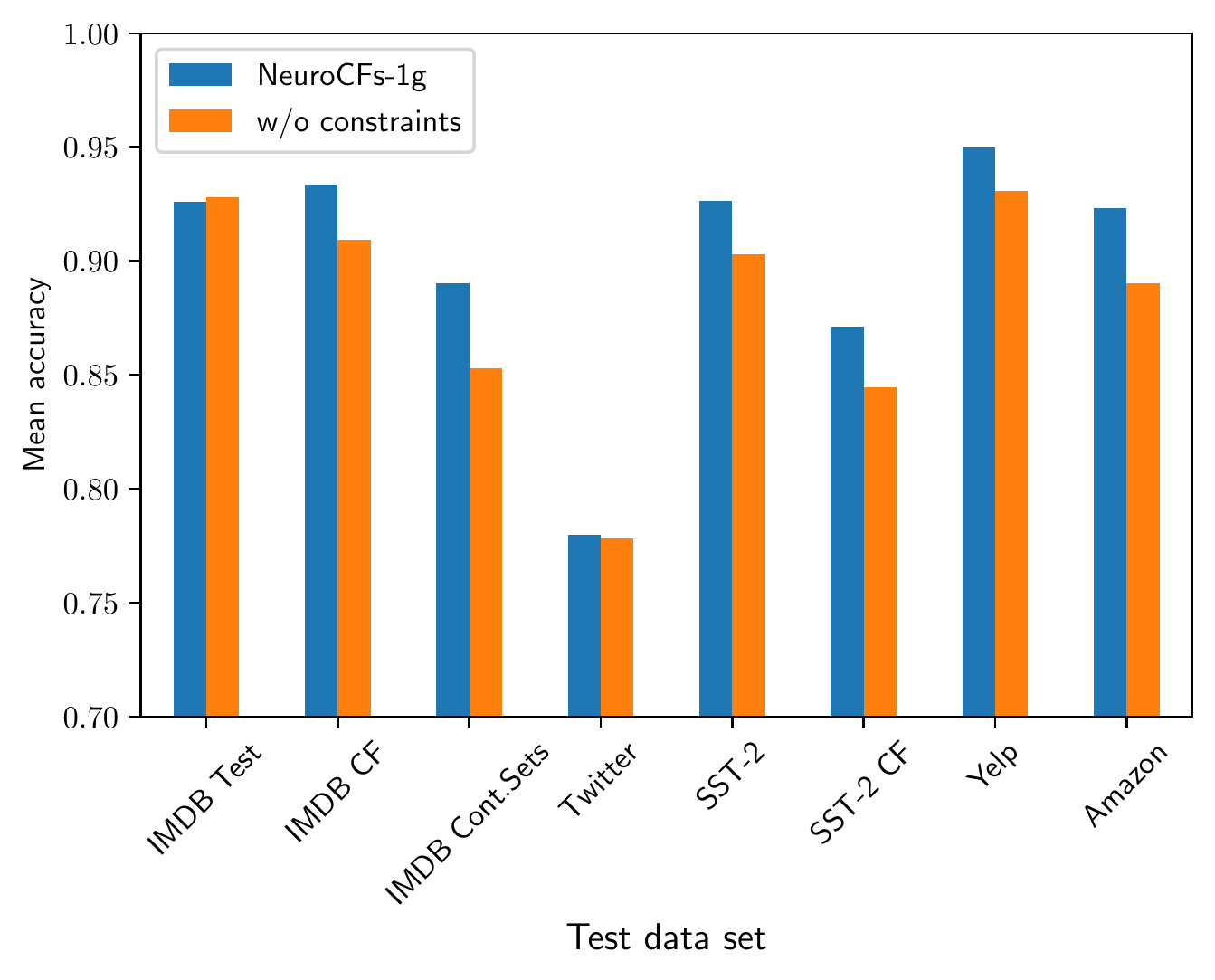}
         \caption{
         Conceptual constraint-based decoding with \neurologic improves performance, as seen by the comparison between training data augmentation with \ncfneu, and their counterparts generated without any constraints. 
         Reported RoBERTa-base accuracy is averaged over 30 random seeds.}
         \label{fig:noConstraints}
\end{figure}


Our approach uses a sentiment-steered LM to control the sentiment of the \ncfsmini, and constraint-based decoding to encourage its similarity to the original example. 
To investigate the impact of constraint decoding, we run an ablation without the use of \neurologic, i.e., only using the sentiment steer. 
Specifically, we use the first token of each original sentence as a prompt and decode from our sentiment experts using beam search with the same hyperparameters as \ncfsmini.

Table~\ref{tab:intrinsicEvaluation} compares both variants by their similarity to the original, and Figure~\ref{fig:noConstraints} compares the performance of training data augmentation with both variants.
The use of constrained-based decoding results in substantial performance improvements over the no-constraint baseline across all evaluation sets except the in-domain IMDB test set. 
This highlights the value of using constraints to encourage similarity to the original, thus resulting in a \ncf, as opposed to simply a new example of the opposite polarity.
These results, along with those from \S\ref{sec:diversity} indicate the existence of an optimal degree of similarity, which is not as high as minimal edit counterfactuals, and not as low as constraint-free counterexamples.

Initial experiments further point to the value of ConceptNet constraints, as opposed to nominal constraints; the former results in more similar \ncfs (see App.~\S\ref{app:np-constraints} for details).

\subsection{Leveraging ConceptNet for alternative constraint sets}
\label{sec:alternative-constraints}

\begin{table*}[ht!]
\centering
\resizebox{1\textwidth}{!}{%
\begin{tabular}{l c c c c c c c c}
\toprule
& \multicolumn{3}{c}{\textbf{IMDB}} & \multicolumn{2}{c}{\textbf{SST-2}} & \multicolumn{3}{c}{\textbf{Out-of-domain}}  \\
\cmidrule(lr){2-4}
\cmidrule(lr){5-6}
\cmidrule(lr){7-9}
\textbf{\ncfsmini Constraints} & \textbf{Test} & \textbf{CF (K. et al.)} & \textbf{Cont.Sets} &\textbf{Test} & \textbf{CF (PolyJuice)} & \textbf{Twitter} & \textbf{Yelp} & \textbf{Amazon} \\
\midrule
Original & $92.66_{0.46}$ & $95.03_{0.47}$ & $90.85_{0.84}$ & $\mathbf{92.27}_{0.39}$ & $\mathbf{88.35}_{0.41}$ & $\mathbf{78.80}_{1.22}$ & $\mathbf{94.51}_{0.87}$ & $\mathbf{92.24}_{0.71}$ \\
Concept-altered & $\mathbf{92.85}_{0.63}$ & $95.26_{0.78}$ & $90.55_{1.10}$ & $91.94_{0.49}$ & $87.98_{0.65}$ & $78.53_{1.51}$ & $94.32_{0.80}$ & $92.01_{0.71}$ \\
Original + concept-altered & $91.83_{0.65}$ & $\mathbf{96.04}_{0.46}$ & $\mathbf{91.86}_{1.06}$ & $91.38_{0.42}$ & $\mathbf{88.35}_{0.43}$ & $78.57_{1.63}$ & $93.90_{0.92}$ & $92.03_{0.63}$ \\
\bottomrule
\end{tabular}}
\caption{
Impact of concept-altered constraint sets created from ConceptNet on classifier performance
}
\label{tab:conceptnetAblation}
\end{table*}

Our use of COCO-EX for identifying concept constraint sets provides a link between each of our constraints and a node in ConceptNet. 
We explore whether the structured knowledge contained in ConceptNet can provide alternative constraint sets for \ncfsmini. 

For each concept in our original constraint sets, we query ConceptNet for its most similar\footnote{
Via similarity scores calculated over pre-computed ConceptNet Numberbatch embeddings.
}
English-language node in the graph and use the label of this nearest neighbor to replace our original concept constraint (see Appendix~\ref{app:conceptAltConstraints} for examples). 
Table~\ref{tab:conceptnetAblation} compares the performance of a RoBERTa-base classifier trained on \ncfneu, and their counterparts produced by alternative conceptual constraints derived from ConceptNet, and a combined set of \ncfsmini produced by both the original and concept-altered constraints. 
We observe that further increasing the size of our CFs using concept-altered \ncfsmini increases performance on in-domain CF test sets while retaining performance on OOD test sets. 
While this pilot shows promising results, we leave a systematic investigation into ConceptNet knowledge to create counterexamples for data augmentation, to future work.

\subsection{Can \ncfsmini be used for evaluation?}
\label{sec:contrast-sets}

Inspired by the success of \ncfs for training data augmentation, we further investigate if these can be used as a challenge set for evaluation \cite{rudinger-etal-2018-gender}.
However,  before deploying them as test sets, we need to first verify that \ncfs indeed alter the target label, as intended by the sentiment steering process (\S\ref{sec:lm-steering}).
We randomly select 50 \ncfs, as well as CFs from baseline approaches, to evaluate whether they successfully steered the sentiment of the original example.\footnote{To ensure fairness, the source of the counterfactual as well as the intended label was kept hidden during validation.}
Results show that \ncfneu and \ncfuni are more successful in steering sentiment compared to the baseline approaches; however, only about 50\% of the resulting \ncfneu actually result in sentiment change; see further discussion in App.~\ref{app:evaluation}.
Hence, we cannot reliably use generated counterfactuals for evaluation.
Future work might investigate manually labeling \ncfs for use as challenge sets, following \citet{wu2021polyjuice}.




\section{Related Work}
\label{sec:related}

Counterfactual data augmentation is emerging as a viable solution for improving model robustness towards spurious correlations \cite{geirhos2020shortcut}.
In previous sections, we present comparisons to various minimal edit approaches for producing counterfactuals \cite{kaushik2019learning,wang2021robustness,yang2021exploring,wu2021polyjuice,gardner2020evaluating}, either manually or automatically.
Our approach steers away from minimal edits, as well as manual intervention for creating counterfactuals.

Beyond sentiment classification, this approach has been employed for tasks such as question answering \cite{paranjape2022retrieval}, fairness in social computing \cite{sen2021counterfactually}, and natural language inference \cite{glockner2018breaking}.
Most work focus on minimal edits of training instances via small perturbations to the causal features, via manually editing instances.
\citet{madaan2021generate} introduce a controlled text generation approach to create counterfactuals containing specific attributes, but focus on applications to debiasing and evaluation rather than our objective of training data augmentation. 
\citet{hu2021causal} propose a structural causal model for combing attribute-conditional generation and text attribute transfer (i.e., minimal edits), but similarly produce counterfactuals for different purposes than ours.
\citet{ross-etal-2022-tailor} automate contrast sets \cite{gardner2020evaluating} for question answering, dependency parsing and relation extraction, via training a generator with semantic control codes; however, their method requires the user to specify what changes in the original sentence are desired.



\swabha{todo: cite \cite{kaushik2021explaining}}

\paragraph{Beyond Counterfactuals:}
\citet{srivastava2020robustness} collect human annotations for common-sense reasoning behind examples, in a robust optimization setting to minimize worst-case loss, without explicitly collecting counterfactuals. 
\citet{ribeiro2018semantically} demonstrate how state-of-the-art models are vulnerable to semantically-equivalent adversarial examples constructed from a rule-based method. 
\citet{ribeiro2020beyond} propose Checklists, which contain heuristic edits of the evaluation data instances.
Other approaches employ perturbations without creating actual data instances \cite{veitch2021counterfactual}.




\section{Discussion}
\label{sec:discussion}

We presented an approach to generate \ncfs, via sentiment steering and concept-constrained decoding.
Training data augmentation with \ncfs results in improvement on sentiment classification performance over existing minimal-edit methods, both in and out of domain; even matching human counterfactuals in some cases.
We presented several analyses for \ncfs, and ablations showing the effectiveness of our approach.
While \ncfs are loose by design, our analyses indicate the existence of an optimal degree of similarity, which is not as high as minimal edit counterfactuals, and not as low as constraint-free counterexamples.

While this work focused on \ncfs for movie reviews only, our results show that training on them transfers to other domains such as product reviews and social media posts for the same sentiment analysis task.
Future directions of research might investigate generating \ncfs for evaluation, and tasks beyond sentiment classification. 
Our approach is broadly compatible with tasks for which a language model steer can be trained; future applications of this work could therefore include other NLP tasks where global attributes are available, such as toxicity removal or style transfer. 
\phillip{Any other ideas for future work?}\swabha{added}
Further, we could consider generating a \ncfs neighborhood around individual instances, similar to contrast sets \cite{gardner2020evaluating}.

\swabha{also see appendix comments}

\section*{Acknowledgments}

We thank Ximing Lu and Chandra Bhagavatula from the Allen Institute for AI for help with the {\neurologic decoding codebase},\footnote{\url{{https://github.com/GXimingLu/neurologic_decoding}}} and members of UW NLP, particularly Alisa Liu and Suchin Gururangan for valuable feedback on earlier drafts of this paper. 
Additionally, we thank Zev Rivlin and Joscha Bach of Intel Labs for their insights throughout the project. 
We would also like to thank the anonymous reviewers for their constructive input.

\section*{Limitations}
\label{sec:limitations}

Our approach to generate \ncfs is designed specifically for binary sentiment classification in English language only.
For generating \ncfsmini, we needed the knowledge of the original example's sentiment polarity; however, it is possible to produce \ncfsmini for both polarities without knowledge of the original label.
Applications to other classification settings might involve the need to train multiple language model steers, which can be challenging in the absence of global labels (for e.g. instance-specific labels in multiple-choice question answering).
\ncfs might need to be filtered for grammaticality and for steering accuracy for their use beyond training data augmentation.
Our approach investigated producing loose counterfactuals at the sentence level; efficient extensions of our approach to paragraph-level transformations were not explored in this work.
Throughout this work, we use RoBERTa-base and GPT2-Large architectures; however, there are more powerful architectures which could potentially improve our results.

It is possible that language generated through automatic approaches, and labeled automatically might contain their own annotation artifacts \cite{gururangan-etal-2018-annotation}, leading to a different set of spurious biases.
Potential harms of generated language include harmful social biases \cite{bender2021parrots}, which were not investigated in this work.
Approaches that involve a human validation phase after data collection \cite{liu2022wanli}, might be explored in future work to mitigate such harms.

\section*{Ethical Considerations}
\label{sec:ethics}

We acknowledge that generated language is susceptible to harmful social biases \cite{bender2021parrots} and toxicity \cite{gehman-etal-2020-realtoxicityprompts}.
We caution practitioners against training models \textit{solely} on model generated data.
We do not filter our training data or our generations for toxicity, bias, or offensiveness.
Hence, we recommend practitioners interested in using our generations and replicating this work to carefully check the generated content before deployment in any real world application. 

Our work uses only publicly available datasets. 
To the best of our knowledge, these do not contain any explicit information about a user's identity, health, negative financial status, racial or ethnic origin, religious or philosophical affiliation or beliefs, beyond their reviews on movies and products.

\bibliography{anthology,custom}
\bibliographystyle{acl_natbib}
\clearpage

\appendix
\section{Extended Qualitative Analysis}
\label{app:more-qualitative}
A larger qualitative analysis is provided in \autoref{tab:qualitative-more}, which further highlights how \ncfs result in more complex changes to the original sentence, and are more successful in sentiment steering than minimal-edit counterfactuals. Minimal edits are at times unable to result in meaningful sentiment flips, and result in reduced grammaticality and pragmatics, producing phrases such as “racism was best” (W\&C), and “part in the game” (Y et.al.).

\subsection{Examples of cases where a counterfactual was not generated}

Table~\ref{tab:noCfExamples} provides examples of sentences for which a \ncfneu was not generated. In these cases, no prefix of the original sentence at least 4 tokens in length was predicted to be neutral. This can be attributed to sentiment-bearing words being present at the start of the sentence. 

\section{Data Augmentation Experimental Setup}
\label{app:experimental}


\subsection{Sentence-level IMDB}
\label{sec:imdb-s}

We augment the training data of a sentence-level version of this dataset (\textbf{IMDB-S}), introduced by \citet{wang2021robustness}.
Here, the original paragraph-length examples were disaggregated, by splitting the original paragraph into sentences and selecting those which contain keywords highly correlated with labels predicted by a binary sentiment classifier. 
Each sentence inherits its label from the original paragraph, and \citet{wang2021robustness} found that $96.8\%$ of the inherited labels were accurate based on a manual evaluation of 500 samples.

\subsection{Dataset Sizes}
\label{sec:dataset-details}

\begin{table}[ht!]
\centering
\small
\begin{tabular}{l c c}
\toprule
\textbf{Dataset} & \textbf{$|D_\textrm{train}|$} & \textbf{$|D_\textrm{test}|$}\\
\midrule
IMDB-S & 8173 & 2245 \\
IMDB-S CF & --- & 2381 \\
IMDB & --- & 488 \\
IMDB CF \cite{kaushik2019learning} & --- & 488 \\
IMDB Cont.Sets \cite{gardner2020evaluating} & --- & 488 \\
SST-2 & --- & 1821 \\
SST-2 CF \cite{wu2021polyjuice} & --- & 3014 \\
Twitter & --- & 4678 \\
Yelp & --- & 38000 \\
Amazon & --- & 941534 \\
\bottomrule
\end{tabular}
\caption{
Size of datasets used in experiments
}
\label{tab:dataSize}
\end{table}

Table~\ref{tab:dataSize} provides details on the size of the datasets used in our experiments. All datasets consist of English language text which we used without modification. 
For training our baselines, \citet{wang2021robustness} provided the sentence-level variants for \citet{kaushik2019learning}'s counterfactuals, and we apply their method to obtain the sentence-level counterfactuals from \citet{yang2021exploring}.

\subsection{Models and Hardware Details}
\label{sec:model-details}

Our sentiment classifier consists of a RoBERTa-base model \cite{liu2019roberta} finetuned on various training data setups for a maximum of $10k$ steps using the AdamW optimizer \cite{Loshchilov2019DecoupledWD} with a batch size of 16 and a learning rate of 1e-06.
We evaluate performance every 500 steps on a validation set randomly sampled from $20\%$ of the training data and terminate training early if there is no improvement for 5 consecutive evaluations. 
All sources of counterfactuals are evaluated using the same hyperparameters and strategy for withholding validation data.

Our experiments were conducted on a Slurm linux cluster with Nvidia RTX 3090 GPUs. We parallelized the generation of \ncfs across 32 GPUs in this environment, resulting in a total running time of 75 minutes. Table~\ref{tab:robertaTrainTime} reports the mean time to train our RoBERTa-base classifier on the various sets of counterfactuals, measured across 30 different random seeds. Each training run for a given source of counterfactuals and seed was conducted on a single GPU. RoBERTa-base has 125M parameters. 

\begin{table}
\centering
\small
\resizebox{0.45\textwidth}{!}{%
\begin{tabular}{l r c}
\toprule
\textbf{Source of CFs} & \textbf{$|D_\textrm{train}|$} &  \textbf{Mean Training Time}\\
\midrule
None & 8,173 & 641.90\\
\midrule
\citealp{yang2021exploring} & 10,376 & 773.79 \\
\citealp{wang2021robustness} & 10,744 & 827.94\\
\ncfneu & 12,905 & 746.86\\
\ncfuni & 15,437 & 927.50\\
\midrule
$\dagger$ \citealp{kaushik2019learning} & 16,679 & 788.13\\
\bottomrule
\end{tabular}}
\caption{
Average time (in seconds) to train RoBERTa-base on various sets of counterfactuals measured across 30 random seeds}
\label{tab:robertaTrainTime}
\end{table}

\begin{table*}[ht]
\small
\resizebox{\textwidth}{!}{
\begin{tabular}{l l}
\toprule
\textbf{Candidate prompt} & \textbf{Original sentence} \\ 
\midrule
\post{Long , boring ,} & \post{Long, boring, blasphemous.} \\
\post{Do something worthwhile ,} & \post{Do something worthwhile, anything really.} \\
\post{Awful , despicable ,} & \post{Awful, despicable, unpleasant, unhappy, unredeemable saga of a complete Loser.} \\
\post{This is a good} & \post{This is a good, dark film that I highly recommend.} \\
\post{I really liked the} & \post{I really liked the black and white cinematography.} \\
\bottomrule
\end{tabular}}
\caption{
Examples of cases where a \ncfneu was not generated
}
\label{tab:noCfExamples}
\end{table*}
\section{Additional Results}
\label{app:more-results}

\begin{table}[ht]
\centering
\small
\begin{tabular}{l r}
\toprule
&  \textbf{Steering Acc.} \\
\midrule
\citealp{yang2021exploring} & 0.24 \\
\citealp{wang2021robustness} & 0.28 \\
\ncfuni & 0.40\\
\ncfneu & 0.46 \\
\bottomrule
\end{tabular}
\caption{
Accuracy of sentiment steering, based on manual evaluation by authors of this work, on 50 randomly sampled IMDB-S train instances for which CFs were available from all approaches.
Many generations from each approach were ungrammatical and unpragmatic (see examples in \autoref{tab:qualitative} and  \autoref{tab:qualitative-more}), and we considered them as incorrectly sentiment-steered.
}
\label{tab:steerAcc}
\end{table}
\begin{table}[ht]
\centering
\small
\begin{tabular}{l r r r}
\toprule
\textbf{Source of CFs}&  $|D_\textrm{test}^\textrm{CF}|$ & \textbf{\begin{tabular}[r]{@{}l@{}}Acc.\end{tabular}} & $\Delta (\uparrow)$\\
\midrule
None & 2245 & 80.46 & 0.0\\
 \midrule
\HandPencilLeft~\citealp{gardner2020evaluating} & 4545 & 67.52 & 12.68\\
\HandPencilLeft~\citealp{kaushik2019learning} & 2381 & 77.57 & 2.63\\
\midrule
\ncfuni & 2051 & 67.63 & 12.57\\
\ncfneu & 1322 & 56.81 & \textbf{25.39} \\
\bottomrule
\end{tabular}
\caption{
Classification accuracy of an off-the-shelf sentiment classifier from the Huggingface Transformers library (RoBERTa-base finetuned on the Yelp dataset). 
Each row indicates an evaluation set comprised of counterfactuals of the original IMDB-S test set (top row), from different sources.
$|D_\textrm{test}^\textrm{CF}|$ indicates size of the counterfactual test set.
\HandPencilLeft~indicates manually created counterfactuals.
Greater the $\Delta$, more challenging the CF test set.
However, \ncfuni and \ncfneu do not possess human-annotated target labels; also see \S\ref{sec:contrast-sets}.
}
\label{tab:testOnCtf}
\end{table}

\subsection{Evaluating on sentence-level test sets}
\label{app:slevel-eval}

\begin{table}[ht!]
\centering
\resizebox{0.45\textwidth}{!}{%
\begin{tabular}{l r c c}
\toprule
& & \multicolumn{2}{c}{\textbf{IMDB-S}} \\
\cmidrule(lr){3-4}
\textbf{Source of CFs} & \textbf{$|D_\textrm{train}|$} &  \textbf{Test} & \textbf{CF (K. et al.)}\\
\midrule
None & 8,173 & $80.46_{0.55}$ & $75.21_{0.84}$ \\
\midrule
\citealp{yang2021exploring} & 10,376 & $79.68_{0.50}$ & $77.63_{0.71}$\\
\citealp{wang2021robustness} & 10,744 & $\mathbf{80.25}_{0.42}$ & $77.62_{0.90}$\\
\ncfneu & 12,905 & $78.31_{0.53}$ & $\mathbf{80.01}_{0.71}$\\
\ncfuni & 15,437 & $79.03_{0.56}$ & $77.87_{0.77}$\\
\midrule
\HandPencilLeft~\citealp{kaushik2019learning} & 16,679 & $77.58_{0.39}$ & $\mathbf{84.27_{0.46}}$\\
\bottomrule
\end{tabular}}
\caption{
Evaluation of counterfactual data augmentation on sentence-level test sets; other settings similar to \autoref{tab:fullCtfSet}.
}
\label{tab:imdb-s-test}
\end{table}
\autoref{tab:imdb-s-test} shows the results of all our approaches and baselines on sentence-level test sets.

\subsection{Noun Chunk Concepts as Constraints}
\label{app:np-constraints}

As detailed in Section~\ref{sec:constraints}, we form our constraint sets by using COCO-EX to identify meaningful concepts in the original example. 
To investigate how our concept-constrained generations differ from those produced by constraint sets derived from nouns, we generated an alternative set of \ncfs using constraints consisting of noun chunks identified by spaCy\footnote{\url{https://spacy.io/}}. 
We found that these alternative noun-chunk conceptual \ncfs had an average MoverScore of 0.15 w.r.t. their corresponding COCO-EX concept-constrained \ncfs, indicating that the use of concepts for constraint formulation produces substantially different counterfactuals than the use of noun chunks for constraints. 
Moreover, based on the evidence from \autoref{tab:ctfSimilarity}, we hypothesize that these alternative concepts might not result in a performance boost.

\subsection{Examples of concept-altered constraint sets derived from ConceptNet}
\label{app:conceptAltConstraints}

\begin{table*}[ht]
\small
\resizebox{\textwidth}{!}{
\begin{tabular}{p{5cm} p{2.8cm} p{2.5cm} p{5cm} p{5cm}}
\toprule
\textbf{Original Sentence} & \textbf{Original Constraints} & \textbf{Constraints w/ Altered Concepts} & \textbf{\ncfneu} & \textbf{Concept-altered \ncfneu}\\ 
\midrule
\post{\textcolor{Orange}{This is one of the} worst movies I saw!} & \post{(movies)} & \post{(citizen kane)} & \post{This is one of the funniest movies I have seen in a long time.} & \post{This is one of the best movies I've seen in a long time, and it's also a movie that will make you laugh, cry, think and feel a little bit like citizen kane.}\\
\midrule
\post{\textcolor{Orange}{It's maybe the} worst comedy spoof ever made.} & \post{(spoof) $\land$ (comedy)} & \post{(parodied) $\land$ (comedic)} & \post{It's maybe the best spoof comedy I've seen in a long time.} & \post{It's maybe the most parodied comedic film I've seen in a long time.}\\
\midrule
\post{\textcolor{Orange}{Unlike many modern stories which seem to revel in dark witchcraft, this is simply} a magical tale of hocus pocus that is cute, light hearted, and charming.} & \post{(hocus pocus) $\land$ (tale)}  & \post{(mumbo jumbo) $\land$ (story)}  & \post{Unlike many modern stories which seem to revel in dark witchcraft this is simply a tale of hocus pocus and sleight of hand.} & \post{Unlike many modern stories which seem to revel in dark witchcraft this is simply a story about mumbo jumbo and a lot of it.}  \\
\midrule
\post{\textcolor{Orange}{He really just wants to be a good boy, to do the right thing}, and to make his brother proud of him.} & \post{(brother)} & \post{(younger sibling)} & \post{He really just wants to be a good boy to do the right thing for his brother, but he just can't do it.} & \post{He really just wants to be a good boy to do the right thing, but his younger sibling isn't buying it.} \\
\bottomrule
\end{tabular}}
\caption{
Examples from IMDB-S and their corresponding \ncfneu, generated with original and with concept-altered constraints (see \S\ref{sec:alternative-constraints}).
The prompt (history) used for \neurologic decoding is colored \textcolor{Orange}{orange}.
}
\label{tab:conceptAlteredExamples}
\end{table*}

Table~\ref{tab:conceptAlteredExamples} provides examples of our original \ncfneu and their concept-altered versions after replacing constraints with similar nodes from ConceptNet.


\subsection{Evaluating with \ncfs}
\label{app:evaluation}

\autoref{tab:steerAcc} shows the steering accuracy of \ncfs as well as CFs from baseline approaches, as evaluated by the authors of this work on a sample of 50 randomly selected examples from each.
Some examples of this annotation can be seen in Table~\ref{tab:qualitative-more} in Appendix~\ref{app:more-qualitative} and in Table~\ref{tab:qualitative}.

We report the performance of a RoBERTa-base classifier finetuned on the Yelp dataset\footnote{\url{https://huggingface.co/VictorSanh/roberta-base-finetuned-yelp-polarity}} using the original IMDB dataset and various CF test sets in Table~\ref{tab:testOnCtf}.

\begin{table*}[ht]
\small
\resizebox{\textwidth}{!}{
\begin{tabular}{@{}p{1.9cm}p{0.5cm}p{16.5cm}@{}}
\toprule
& \textbf{Label} & \textbf{Review} \\ 
\midrule
Original & \positive 
& \post{A good enough film that unfortunately leaves you a little sad at the end.} \\
W\&C. &  \positive
& \post{A good enough film that luckily leaves you a little sad at the end} \\
Y.et al. & \positive
& \post{A good enough film unfortunately leaves you a little sad at the end.}  \\
\ncfuni &  \neutral
& \post{A film about the end of the world as we know it.}  \\
\ncfneu & \negative
& \post{A good enough film that unfortunately leaves you a little sad at the end, but it's not a great one.}  \\
\midrule
Original & \positive & \post{Crash tried to show how racism was bad (and Crash actually had a built-in anti Asian bias) and to come at it from a morally superior position.}\\
W\&C. & \negative & \post{Crash tried to show how racism was best and crash actually had a built in anti asian bias and to come at it from a morally superior position} \\
Y.et al. & \neutral & \post{Crash tried to show how was bad ( and crash actually had a built - in anti asian bias ) and to come at it from a morally superior position.} \\
\ncfuni & \neutral & \post{Crash crash, anti racism position, asian stereotypes, and.} \\
\ncfneu & \negative & \post{Crash tried to show how racism was bad and crash actually had a built-in anti asian bias and to come at it from a morally superior position, it didn't do much of either.} \\
\midrule
Original & \positive & \post{Many references to the directors obvious love of film history (a great "Bicycle Thief" homage") and some whimsical scenes which work with out being pretentious.}\\
W\&C. & \positive & \post{Many references to the directors obvious love of film history a little bicycle thief homage and some whimsical scenes which work with out being pretentious} \\
Y.et al. & \positive & \post{Many references to the directors obvious love of film history ( a great " bicycle thief " homage " ) and some whimsical which work with out being pretentious.} \\
\ncfuni & \negative & \post{Many history buffs will love this film, but it's not for directors or actors, and it doesn't have the charm of scenes like this one, in which a bicycle thief references the film.} \\
\ncfneu & \negative & \post{Many references to the directors obvious love of film history and scenes from the classics are made, but the film is so formulaic that it's hard to care about the characters or the bicycle thief.} \\
\midrule
Original & \negative & \post{The motorcycle is the worst part in the game.} \\
W\&C. & \positive & \post{The motorcycle is the good part in the game} \\
Y.et al. & \neutral & \post{The motorcycle is the part in the game.} \\
\ncfuni & \positive & \post{The best part of the game is the motorcycle chase sequence.} \\
\ncfneu & \positive & \post{The motorcycle is the best part of the film, and it's a game changer.} \\
\midrule
Original & \negative & \post{Where this weakness is most apparent is in his humor, which is poised and logical, so the joke is dead before it's delivered.} \\
W\&C. & \negative & \post{Where this weakness is most apparent is in his humor which is poised and logical so the well is dead before it s delivered} \\
Y.et al. & \negative & \post{Where this is most apparent is in his humor, which is poised and logical, so the joke is dead before it's delivered.} \\
\ncfuni & \positive & \post{Where the joke is, the humor comes from the characters.} \\
\ncfneu & \negative & \post{Where this weakness is most apparent is in his humor which is poised and logical so the audience doesn't have to guess what the joke is.} \\
\midrule
Original & \positive & \post{Jacknife is never getting boring as it shows its heroes clear of any typical Hollywood's typical character elements.} \\
W\&C. & \negative & \post{Jacknife is never getting interesting as it shows its heroes clear of any typical hollywood s typical character elements} \\
Y.et al. & \negative & \post{Jacknife is getting boring as it shows its heroes clear of any typical hollywood's typical character elements.} \\
\ncfuni & \negative & \post{Jacknife offers elements of hollywood melodrama, character development and heroes and villains.} \\
\ncfneu & \negative & \post{Jacknife is never getting the character elements hollywood heroes are supposed to have.} \\
\midrule
Original & \negative 
& \post{The scenery looks like cheap Theatre.} \\
W\&C. & \positive
& \post{The scenery looks like \underline{expensive} theatre}  \\
Y.et al. & \neutral
& \post{The scenery looks like \sout{cheap} theatre.}  \\
\ncfuni &  \positive
& \post{The scenery, the music, and the theatre are all top notch.}  \\
\ncfneu & \neutral 
& \post{The scenery looks like it could have come straight out of a movie theatre.}  \\
\bottomrule
\end{tabular}}
\caption{
\swabha{Add blue highlights to all sentences.}
Further qualitative analysis, extending \autoref{tab:qualitative}.
}
\label{tab:qualitative-more}
\end{table*}

\end{document}